\documentclass[letterpaper, 10 pt, conference]{ieeeconf} 

\IEEEoverridecommandlockouts                       
\usepackage{tikz}
\usetikzlibrary{shapes.geometric, positioning, calc}
\usepackage{graphicx}
\usepackage{amsmath} 
\usepackage{amssymb}
\usepackage{cite}

\usepackage{tabularx}
\usepackage{array}
\usepackage{booktabs}
\usepackage{bm}
\usepackage{algorithm}
\usepackage{algorithmic}
\usepackage{bbm}
\usepackage{caption}
\PassOptionsToPackage{hyphens}{url}\usepackage{hyperref}

\usepackage{fancybox}        
\usepackage{varwidth}        

\usepackage{mathtools}       
\usepackage{xparse}          
\usepackage{multirow}        
\usepackage{boxedminipage}   
\usepackage{flushend}
\usepackage{soul}

\definecolor{myYellow}{RGB}{255, 255, 0}
\definecolor{myBlue}{RGB}{69, 114, 196}

\definecolor{green}{RGB}{11,155,13}

\title{\LARGE \bf

Social-LLaVA: Enhancing Robot Navigation through Human-Language Reasoning in Social Spaces

}

\author{Amirreza Payandeh$^1$, Daeun Song$^1$, Mohammad Nazeri$^1$, Jing Liang$^2$, Praneel Mukherjee, \\ Amir Hossain Raj$^1$, Yangzhe Kong$^1$, Dinesh Manocha$^2$, and Xuesu Xiao$^1$
\thanks{$^1$George Mason University {\tt\scriptsize \{apayande,  dsong26, mnazerir, araj20, ykong7,  xiao\}@gmu.edu, praneel.mukherjee@gmail.com} $^2$University of Maryland, College Park {\tt\scriptsize \{jingl, dmanocha\}@umd.edu}}
}

\begin{document}
\maketitle
\thispagestyle{empty}
\pagestyle{empty}

\begin{abstract}

Most existing social robot navigation techniques either leverage hand-crafted rules or human demonstrations to connect robot perception to socially compliant actions. However, there remains a significant gap in effectively translating perception into socially compliant actions, much like how human reasoning naturally occurs in dynamic environments. Considering the recent success of Vision-Language Models (VLMs), we propose using language to bridge the gap in human-like reasoning between perception and socially aware robot actions. We create a vision-language dataset, Social robot Navigation via Explainable Interactions (\textsc{snei}), featuring 40K human-annotated Visual Question Answers (VQAs) based on 2K human-robot social interactions in unstructured, crowded public spaces, spanning perception, prediction, chain-of-thought reasoning, action, and explanation. We fine-tune a VLM, Social-LLaVA, using \textsc{snei} to demonstrate the practical application of our dataset. Social-LLaVA outperforms state-of-the-art models like GPT-4V and Gemini, based on the average of fifteen different human-judge scores across 50 VQAs. Deployed onboard a mobile robot, Social-LLaVA enables human-like reasoning, marking a promising step toward socially compliant robot navigation in dynamic public spaces through language reasoning\footnote{Website: \url{https://cs.gmu.edu/~xiao/Research/SNEI/}}.
\end{abstract}

\section{Introduction}
\label{sec::introduction}

As mobile robots become more prevalent in human-centric environments, there is a growing interest in social navigation, augmenting traditional methods by aligning with human social norms and rules rather than merely treating humans as dynamic or static obstacles~\cite{mirsky2024conflict, francis2023principles, mavrogiannis2023core}. An extensive body of work has addressed social robot navigation, ranging from employing various hand-crafted navigation techniques based on geometric and semantic understanding~\cite{van2011reciprocal, helbing1995social, ferrer2013social} to learning-based methods using large-scale datasets~\cite{scand, musohu, hirose2023sacson}. While these methods have made progress toward achieving socially compliant behaviors, they often fail to grasp the nuances of the context and scene in the same way humans do.
\begin{figure}[!htbp]
    \centering
    \includegraphics[width=0.5\textwidth]{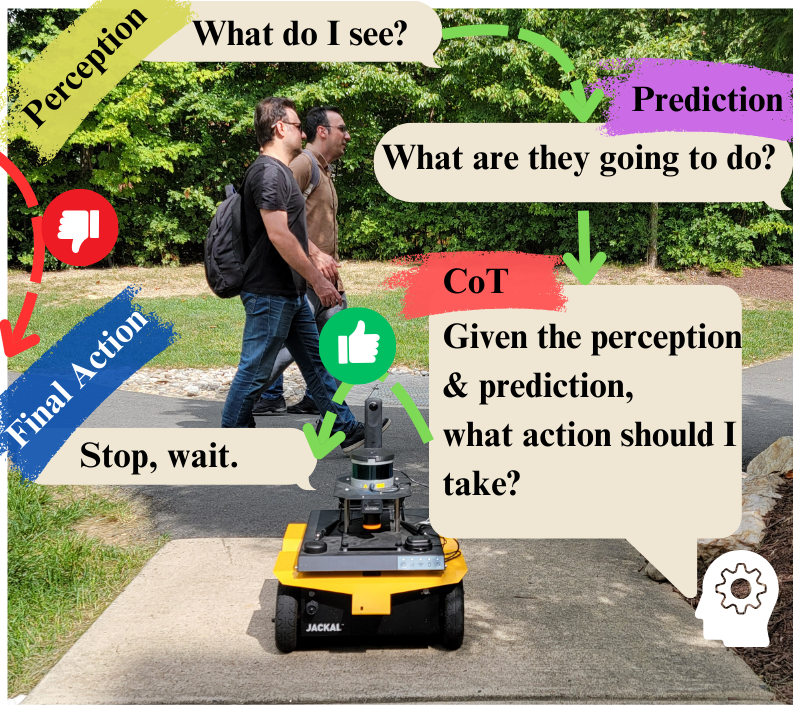}
    \caption{Bridging perception to socially compliant action through Chain-of-Thought reasoning using human language. }
    \label{fig::first}
\end{figure}

\begin{figure*}[!htbp]
    \centering
    \includegraphics[width=\textwidth]{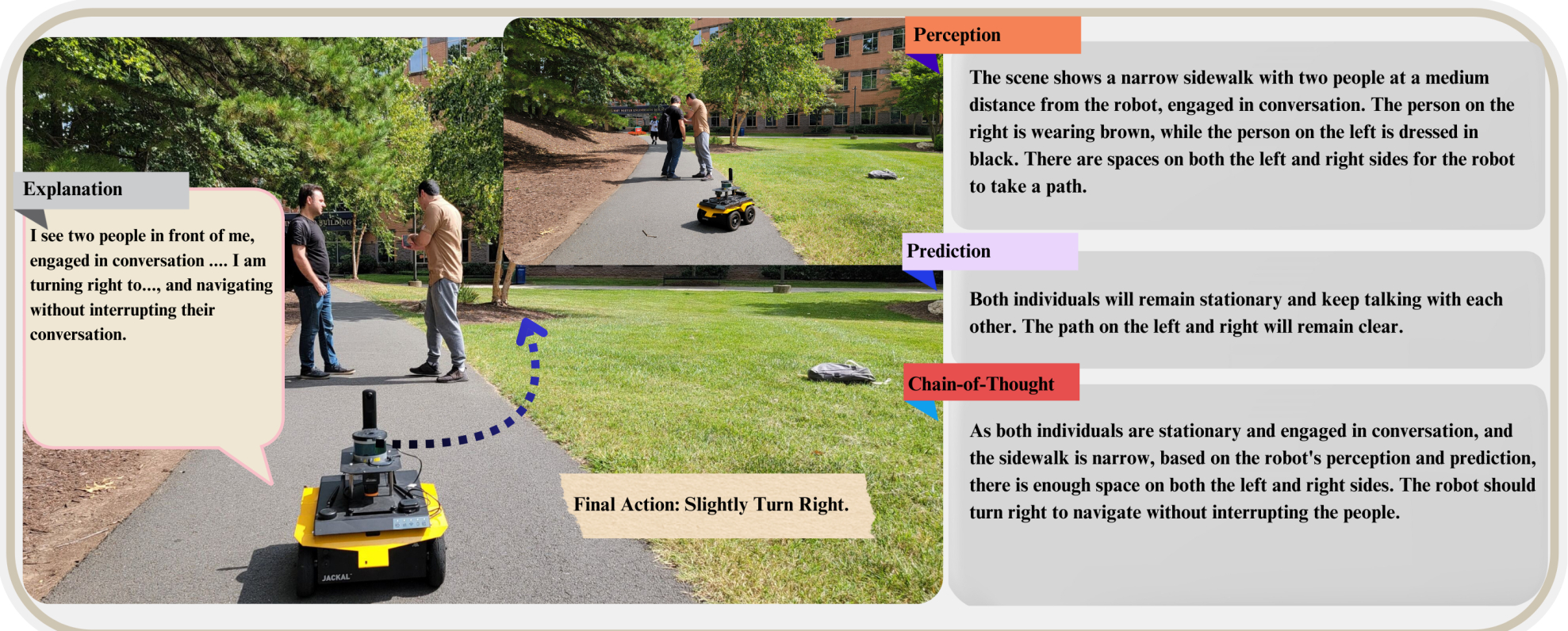}
        \caption{Social-LLaVA:  Proof-of-concept real-world experiment demonstrating the robot's ability to understand context and social cues to navigate, thereby avoiding interrupting people's conversations.
        }
    \label{fig::real-world}
\end{figure*}

Several studies suggest that language significantly influences human cognition, reasoning, and comprehension of the world. We hypothesize that for robots to behave more like humans, their actions can be guided by reasoning that mirrors human-like, language-based thought processes. If robots can perceive a scene, make predictions like humans, engage in reasoning, and generate action descriptions in human language, they are more likely to translate their perception into actions that closely resemble human behaviors Fig.~\ref{fig::first}.

The recent rapid development of Vision-Language Models (VLMs) and their improving reasoning capabilities have catalyzed interest in applying human language to tasks involving both visual and spatial reasoning, such as visual scene understanding and visual navigation. 
However, trained on general image and language datasets, existing VLMs still face challenges in understanding complex social interactions, especially in spatial and temporal commonsense reasoning and high-level decision-making for social robot navigation, which commonly don't present in those datasets.  
While several large-scale datasets have been proposed for autonomous driving in structured road networks with clear traffic rules, they cannot be directly applied to social robot navigation in unstructured, human-crowded spaces, where agents often follow mostly unwritten and subjective social norms.

To bridge this gap, we introduce a human-annotated vision-language dataset for Social robot Navigation via Explainable Interactions (\textsc{snei}). \textsc{snei} includes over 40K manually annotated Visual Question Answering (VQA) pairs that span perception, prediction, chain-of-thought reasoning, action, and explanation, enabling robots to better understand and respond to their surroundings through natural language descriptions of social context, visual cues, and behavioral patterns. \textsc{snei} is based on 2K manually chosen, unique social scenarios, where a robot engages in direct interaction with humans, from \textsc{scand}~\cite{scand}, a social robot navigation demonstration dataset collected in diverse, human-crowded public spaces and containing complex human-robot interaction scenarios. 

We also propose Social-LLaVA, a baseline model that is fine-tuned from LLaVA~\cite{llava} on \textsc{snei}. This adaptation is specifically tailored to our dataset, enabling generation of high-level navigation action instruction through chain-of-thought reasoning in human language. Our analysis shows that our Social-LLaVA model outperforms two state-of-the-art models, GPT-4V and Gemini, based on the average of fifteen different human-judge scores across 50 VQA social navigation tasks.


\section{Related Work}
\label{sec::related_work}
In this section, we review related work in social robot navigation and visual instruction tuning for visual navigation.

\subsection{Social Navigation}
Extensive research has been conducted in the field of social robot navigation, driven by the need for robots to operate safely and efficiently in human-populated environments~\cite{mavrogiannis2023core, mirsky2024conflict, francis2023principles,vanp}. The complexity of social navigation arises from the necessity to consider a wide range of factors, such as safety, comfort, politeness, and adherence to unwritten social norms that humans instinctively follow~\cite{pirk2022protocol}. 
Traditionally, model-based approaches that rely on task-specific, hand-engineered behaviors have been employed in social navigation. One of the earliest models is the Social Force Model (SFM)~\cite{helbing1995social, ferrer2013robot}, which simulates human navigation by modeling forces between individuals and obstacles. Another is human-robot proxemics~\cite{mumm2011human, charalampous2016robot}, which focuses on the spatial distances humans maintain around robots to ensure comfortable interactions. 

To overcome the limitations of hand-engineered features and lack of adaptability, recent research has increasingly turned to learning-based methods, such as Learning from Demonstration (LfD)~\cite{9679193,raj2024targeted, 7539621, hirose2023sacson, xiao2020appld, xiao2022learning,10323465}. These techniques allow robots to learn socially compliant behaviors by observing and replicating human demonstrations. Despite these advances in learning-based methods, simply replicating human trajectory from demonstrations in terms of perception-action pairs without human-like reasoning in between can be overly brittle and may not be sufficient to achieve socially aware navigation in a wide range scenarios.

In this work, we hypothesize that human-like language reasoning and explanation between robot perception and action can facilitate socially compliant navigation behaviors. 
Therefore, we create our \textsc{snei} dataset that incorporates human-like comprehension of social contexts in terms of language, including the ability to perceive the current situation, predict the actions of other agents, and generate socially compliant navigation behaviors through chain-of-thought reasoning. 

\subsection{Visual Instruction Tuning}
Visual instruction tuning is a method of fine-tuning on an image-text dataset that trains the model to follow textual instructions with visual inputs and generate the desired outputs, thereby enhancing zero-shot performance on specific tasks. The success of multimodal models, such as InstructBLIP~\cite{instructBlip} and LLaVA~\cite{llava}, heavily depends on high-quality general visual instruction tuning datasets~\cite{liu2024mmbenchmultimodalmodelallaround, zhao2023svitscalingvisualinstruction}.

Several general-purpose datasets, mostly generated automatically or semi-automatically, have demonstrated improvement on models' performance~\cite{huang2023visual}. LLaVA~\cite{llava} systematically constructed the LLaVAInstruct-150K dataset by prompting GPT-4 to generate questions and answers using image captions and object bounding boxes from the COCO~\cite{cocoDataset} dataset. InstructBLIP~\cite{instructBlip} integrated VQA datasets for academic tasks related to visual comprehension. LAMM~\cite{LAMM} collected images and point clouds from publicly available datasets, using the GPT-API and self-instruction methods to generate instructions. ALLaVA~\cite{ALLaVA} is an open-source dataset for fine-tuning VQA models, with detailed captions, instructions, and GPT-4V-generated question-answer pairs from a single interaction per image.
In Autonomous Driving, DriveLM~\cite{drivelm} constructed a graph VQA dataset for training VLMs for end-to-end driving. LingoQA~\cite{LingoQA} proposed a video QA dataset for autonomous vehicle explainability.
However, many commonly used instruction-tuning datasets have been found to unexpectedly contain a considerable number of low-quality instances, featuring incorrect or irrelevant responses, potentially due to the (semi-)automatic nature of their collection methods. On the other hand, several works have shown that small, high-quality, human-curated datasets can boost model performance compared to large-scale noisy datasets~\cite{lessismore,chen2024alpagasustrainingbetteralpaca,cao2024instructionmininginstructiondata,wei2023instructiongpt4200instructionparadigmfinetuning}. 
Based on these insights, and given that no existing general-purpose datasets adequately address the complexities of human-robot social navigation interactions (refer to our experimental results for details), we propose a novel visual instruction tuning dataset. This dataset features language-based human annotations that encompass perception, prediction, chain-of-thought reasoning, action, and explanation, all tailored for social robot navigation.

\section{The \textsc{snei} Dataset}
\label{sec::dataset}

In this section, we present the motivation of our approach, formulate our social navigation VQA task, and introduce the \textsc{snei} dataset. 

\subsection{Motivation}

To leverage human-language reasoning, robots first need to transform their visual perceptions and predictions derived from those perceptions into language. Then, using established chain-of-thought reasoning methods~\cite{cot}, they can produce high-level actions using human-like language reasoning~\cite{groeger2013understanding}. Inspired by DriveLM~\cite{drivelm}, we hypothesize that mobile robots can leverage a decision-making process that humans implicitly perform, i.e., object-centric perception, prediction, and planning, in the format of language to describe each of these three stages~\cite{marr_vision_2010}.

Our preliminary experiments with off-the-shelf state-of-the-art VLMs reveal significant limitations in spatial reasoning, particularly in tasks critical to social robot navigation, such as determining the relative positions of humans, estimating their intent, and predicting their trajectories.
As highlighted by SpatialVLM~\cite{spatialreasoning}, these shortcomings in spatial reasoning capabilities of state-of-the-art VLMs are attributed more to the limitations of the common datasets used for training than to the models' architectures themselves. This gap can be more pronounced for tasks involving complex, dynamic environments with multiple interacting humans, requiring precise understanding of both spatial relationships and social cues.
The majority of available VQA datasets for visual navigation are either general-purpose or task-specific, such as those for autonomous driving. However, due to the distinct nature of these tasks, such datasets are not directly applicable to mobile robot navigation in human-populated public spaces. Furthermore, the (semi-)automated methods used for their collection often result in a significant amount of noise.

To this end, we propose a dataset of over 40K VQA instances fully annotated by humans for mobile robot navigation in unstructured, crowded environments. To the best of our knowledge, this is the first VQA dataset specifically designed for social robot navigation.

\subsection{Data Construction and Analyses}
We provide two types of annotations: Categorical labels ensure consistency and structure across key elements like crowd density, agent types, and robot actions, while free-form natural language annotations offer greater expressiveness and nuanced descriptions.

\subsubsection{Categorical Labels}
All annotations of this type are selected from predefined categorical options to ensure consistent labeling across the dataset. We use the following categories:  
\begin{itemize}
 \item\textbf{Context} includes environment type,  crowd density, indoor/outdoor status, and terrain type;
 \item\textbf{Robot} includes robot goals, movement directions, speed levels, and action intentions;
 \item\textbf{Obstacle(s)} includes type (e.g., walls and trash cans), proximity, and position relative to the robot; and
 \item\textbf{Agent(s)} includes type (e.g., individuals, groups, and bicycles), proximity, position relative to the robot, current action, and facing direction.
\end{itemize}

\subsubsection{Free-form natural language}
All annotations of this type take the form of natural language descriptions to cover the following aspects: 

\begin{itemize}
\item\textbf{Perception} describes the robot's visual inputs, focusing on humans, including their clothing color, position, relative distance, action, and surrounding crowd density;
\item\textbf{Prediction} assesses the potential future movement of agent(s) within the scene;
\item\textbf{Chain-of-Thought Reasoning} given the perception and prediction, formulates a high-level natural language instruction (e.g., Given the close proximity of the person crossing the robot's path from left to right, the robot should stop, wait for the person to pass, and then continue);
\item\textbf{Final Action} comprises high-level natural language action commands (e.g., Stop and wait for clear path); and
\item\textbf{Explanation} includes a general explanation of what the robot sees, what the robot does, and why.
\end{itemize}

We use the \textsc{scand} dataset~\cite{scand}, which is collected from various human-crowded public environments and features intricate human-robot interaction scenarios. We manually choose and label 2K scenarios where the robot interacts with people.

\begin{figure*}[t]
\centering
\includegraphics[width=\textwidth]{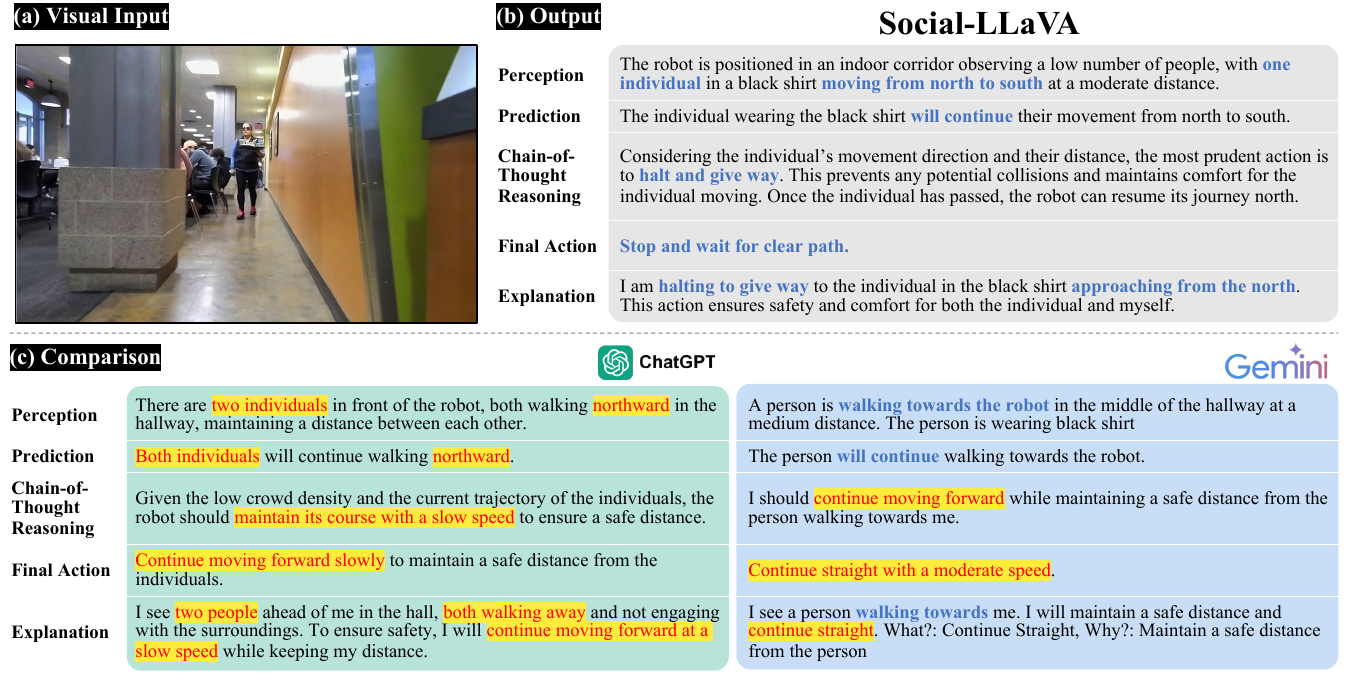}
\caption{Qualitative results of our Social-LLaVA model fine-tuned on our \textsc{snei} dataset compared against GPT4-V~\cite{openai2024gpt4technicalreport} and Gemini 1.5 Pro~\cite{team2023gemini}. (a) shows the visual input given to the models. Note that the given scenario is where a robot navigates through a narrow passage partially obstructed by a pillar (colored red), while an individual approaches the robot (red arrow). (b) illustrates the output from Social-LLaVA, while (c) provides comparisons with GPT4-V and Gemini 1.5 Pro. Phrases in \textcolor{myBlue}{\textbf{blue}} indicate accurate reasoning and socially compliant results, while {\color{red}\hl{highlighted}} phrases mark instances of hallucination.}
\label{fig:comp}
\end{figure*}

\section{\textsc{snei} Applications}
Our \textsc{snei} dataset fills a gap for the social robot navigation community by providing a large-scale, standardized, human-annotated dataset to leverage language to facilitate socially compliant robot navigation. In this section, we show that our dataset can enable human-level language reasoning when facing social navigation interactions in human-crowded public spaces. 

\subsection{Social-LLaVA}
Using our \textsc{snei} dataset, we develop a VLM, Social-LLaVA, which learns to perform human-like reasoning when facing social navigation interactions. 
Notice that our main focus in this work is to provide a standardized dataset specifically designed for mobile robot navigation in human-crowded environments. Therefore, we do not make any claims regarding algorithmic novelty. Instead, we fine-tune an off-the-shelf, state-of-the-art model as a proof-of-concept for the practical application of our \textsc{snei} dataset. 
Social-LLaVA is a fine-tuned version of LLaVA-v1.5-7B~\cite{llava}, which integrates a vision encoder, a large language model, and a vision-language connector. We fine-tune the LLaVA-v1.5-7B on our dataset using Low-rank Adaptation (LoRA)~\cite{lora}, with a batch size of 4 on a single A100 GPU for 15 epochs.

\subsection{Qualitative VQA Evaluation}
To test our Social-LLaVA model fine-tuned on our \textsc{snei} dataset, a large-scale VQA benchmark with a guaranteed performance is required. However, to the best of our knowledge, such a benchmark for social robot navigation does not currently exist. Therefore, we choose to qualitatively analyze the results using human judges. For such a qualitative analysis, we evaluate the free-form natural language outputs generated by Social-LLaVA, compared against two other state-of-the-art foundation models. We prompt Gemini 1.5 Pro~\cite{team2023gemini} and GPT4-V~\cite{openai2024gpt4technicalreport} with extensive explanations of the task and three complicated examples for prompt tuning. The human rater chooses a score between 1-5 for each image-answer, and finally, we average over each task for each model. Table~\ref{tab::scores} shows the achieved scores from the three models for each task. The results demonstrate that Social-LLaVA achieves significantly higher scores compared to GPT4-V and Gemini. In Fig.~\ref{fig:comp} we showcase one of the examples from our dataset and the answers from the three models. 
In this scenario, a robot navigates through a narrow passage partially obstructed by a pillar on the left as an individual approaches from the other side. The results demonstrate that Social-LLaVA successfully generates accurate descriptions for both perception and prediction. Additionally, it provides socially compliant robot actions through effective chain-of-thought reasoning. In contrast, GPT-4V and Gemini 1.5 Pro fail to produce valid responses. Both models incorrectly suggest that the robot should continue moving forward, which would block the individual's path due to the narrow passage. Notably, GPT-4V generates hallucinated outputs in all perception, prediction, chain-of-thought reasoning, final action, and explanation.

This experiment underscores the importance of high-quality VQA data for mobile robot navigation, as it is essential for scene understanding, high-level human trajectory prediction, and chain-of-thought reasoning, which are missing from the pretraining datasets of current state-of-the-art VLMs. We speculate that both aforementioned models would demonstrate significantly improved few-shot performance if fine-tuned on our dataset. However, we cannot validate this speculation, as image fine-tuning is currently unavailable for these models.

\begin{table}[t]
    \centering
    \begin{tabular}{cccc}
        \toprule
         & GPT4-V & Gemini 1.5 Pro & Social-LLaVA \\ 
        \midrule
        Perception & $3.11$ & $3.45$ & $\mathbf{4.0}$\\    
        Prediction & $3.18$ & $3.87$ & $\mathbf{4.06}$\\
        CoT & $3.41$ & $3.79$ & $\mathbf{4.08}$\\
        Final Action & $2.77$ & $3.46$ & $\mathbf{4.19}$\\
        Explanation & $3.16$ & $3.66$ & $\mathbf{3.95}$\\

        \bottomrule
    \end{tabular}
    \caption{A comparative analysis of the performance of the fine-tuned LLaVA model on our dataset against state-of-the-art models, based on average scores per task as evaluated by fifteen human judges. The scores are between 1 to 5.}
    \label{tab::scores}
\end{table}

\subsection{Real World Robot Experiment}
While our \textsc{snei} dataset with the preliminary proof of concept Social-LLaVA model shows potential for enabling socially compliant robot navigation behaviors through explainable interactions using language, producing concrete robot actions based on language descriptions remains an open problem. In this work, we present a simple proof-of-concept using a hard-coded relationship between high-level language descriptions and low-level robot actions (go straight, turn left, and turn right). In Fig. \ref{fig::real-world}, we showcase Social-LLaVA's output after training on our \textsc{snei} dataset in a real-world experiment, where it understands the context, interprets social cues, and avoid interrupting people’s conversations.

\section{Discussions}
\label{sec::discussion}

We discuss our findings through the collection of our \textsc{snei} dataset, development of Social-LLaVA, and evaluation of all three VLMs. 

\subsection{Limited Spatial Understanding}

There are several limitations with current state-of-the-art vision models in spatial understanding, particularly in distinguishing basic spatial concepts like ``left'' and ``right,'' ``above'' and ``below,'' as well as more complex relationships such as ``behind'' and ``in front,'' or relative distances such as ``near'' and ``far''~\cite{spatialgpt,Nejatishahidin2024StructuredSR}. These challenges can be amplified when depth images are not available. Additionally, the complexity and unpredictability of human motion, along with inconsistent behaviors across different individuals, further complicate accurate predictions.

\subsection{Ambiguity in Social Navigation}
The complexity and unpredictability of human motion, along with inconsistent behaviors across different individuals, further complicate accurate predictions and other downstream stages. 
Moreover, the absence of a clear definition of social navigation~\cite{francis2023principles}, which can vary depending on cultural or task-specific contexts, adds to the uncertainty. All these factors make it challenging to generate accurate descriptions and robust predictions for effective decision-making.
These challenges also further complicate our \textsc{snei} data annotation effort, considering that there may be multiple ways or there may not be an agreed-upon way  of socially compliant behavior. 
With these challenges in mind, we aim to generate the most accurate language descriptions possible from a single image to reason based on them and generate high-level actions. While this work serves as a proof-of-concept for using language descriptions as a tool for social robot navigation reasoning, particularly when more than simple obstacle avoidance is required, how to address such ambiguity when annotating data and during training remains open problem. 

\subsection{Need for More and Diverse Data}
While several research efforts show that with a small amount of high-quality data it is possible to fine-tune VLMs, the data-driven nature of these models should not be overlooked~\cite{lessismore,chen2024alpagasustrainingbetteralpaca,cao2024instructionmininginstructiondata,wei2023instructiongpt4200instructionparadigmfinetuning}. We diversify the interactions in our \textsc{snei} dataset as much as possible. However, it remains limited to the scenarios present in the source dataset, \textsc{scand}, which may not capture the full spectrum of human-robot interactions in unstructured environments. Consequently, this limitation could affect the model’s performance when deployed in previously unseen situations or environments, such as in a different country or culture. 
Another aspect is the need of using video data, instead of a single image, to perform language-based reasoning based on sequential information. Considering the difficulty of fine-tuning video-language models~\cite{drivelm}, we currently limit our data annotation to only images. 
We hope our work inspires further exploration in language-driven visual navigation and serves as a foundation for developing techniques to generate more general and high-quality data automatically, potentially enhancing the generalization capabilities of VLMs in diverse real-world contexts.

\subsection{Social Navigation Explainability}
The rapid development of large-scale deep learning models is driving a paradigm shift in the robotics community from rule-based decision-making systems to data-driven, learning-based approaches. However, this shift often comes at the expense of transparency in decision-making. As mobile robots become more prevalent in human-crowded spaces, it is essential to both translate their decision-making into human-readable actions and ensure these decisions are based on human-like reasoning. While several works have addressed explainability of autonomous vehicles to increase transparency, accountability, and trustworthiness of autonomous driving through different methods, there is limited research on mobile robots, especially those operating in human-crowded spaces. In addition to the primary goal of our dataset to enable socially compliant navigation behavior through human-like language reasoning, it can also be used to train models to understand the scene and explain robots' actions to enhance social navigation explainability.

\subsection{Grounding Language to Action}

The recent advancements in VLMs have accelerated progress in exploratory research on a variety of language-driven visual navigation tasks~\cite{Gu_2022}, e.g., autonomous driving~\cite{mao2023gpt}. 
Despite significant efforts and progress, there remains a gap between research and widely adopted real-world applications. 
At a high level, the task of social robot navigation via human-language reasoning can be divided into two subtasks: first, generating accurate language instructions, and second, grounding those instructions into real-world robot actions. Recent research on grounding natural language instructions for motion planning has primarily focused on long-horizon goals rather than short-horizon decision-making ~\cite{quartey2024verifiablyfollowingcomplexrobot}. While the advancement of foundation models demonstrates strong generalizability, their incorporation into short-horizon navigation tasks still presents challenges, such as the lack of embodied experience, hallucination, and the inability to operate in real time~\cite{zhang2024visionandlanguagenavigationtodaytomorrow}.

Our work primarily focuses on the first subtask, specifically on generating high-level human-language actions that promote socially compliant behaviors in various scenarios, such as waiting in line or using off-road paths to avoid interrupting conversations. 

We acknowledge that grounding the generated natural-language high-level actions on the robot in real-world scenarios is still an open problem and is not a trivial task. 

\section{Conclusions}
\label{sec::conclusions}

In this work, we introduce Social robot Navigation via Explainable Interactions (\textsc{snei}), a vision-language dataset specifically designed to bridge the gap between perception and socially compliant actions in crowded public environments through human-like language-based reasoning. We demonstrate that our model, Social-LLaVA, trained on \textsc{snei}, significantly outperforms state-of-the-art systems such as GPT-4V and Gemini in generating socially compliant navigation descriptions. By leveraging language-based reasoning using \textsc{snei} and Social-LLaVA, our preliminary demonstration shows that robots can understand social interactions and navigate unstructured social spaces, while observing human social norms. Our results mark a promising step toward more intuitive and effective social robot navigation in real-world public spaces through explainable human-language reasoning.

\bibliographystyle{IEEEtran}
\bibliography{IEEEabrv,references}
\end{document}